\documentclass[letterpaper, 10 pt, conference]{ieeeconf}
\IEEEoverridecommandlockouts
\usepackage[utf8]{inputenc}
\usepackage[dvipsnames]{xcolor}
\usepackage{graphicx}
\usepackage{multicol}
\usepackage{multirow}
\usepackage{array}
\usepackage{footmisc}
\usepackage{amssymb}
\usepackage{amsmath}
\usepackage{booktabs}
\usepackage{amsfonts}
\let\labelindent\relax
\usepackage{enumitem}

\usepackage{algorithm}
\usepackage{algpseudocode}
\usepackage{tikz}
\usepackage{float}
\usetikzlibrary{math,shapes}
\usepackage{tabularx}
\usepackage{pbox}
\usepackage{ragged2e}
\usepackage{multirow}
\usepackage{tabularx}
\usepackage{balance}
\usepackage{subfig}
\usepackage{relsize}
\usepackage{hyperref}
\usepackage{fancyhdr}

\usepackage{tabularray}
\usepackage{listings}
\mathchardef\mhyphen="2D%

\newcommand{\SimName}{GrowSplat}
\newcommand{\Setup}{Maxi-Marvin}

\newtheorem{Pro}{Problem}

\usepackage{etoolbox}
\makeatletter
\patchcmd{\@makecaption}
  {\scshape}
  {}
  {}
  {}
\makeatletter
\patchcmd{\@makecaption}
  {\\}
  { -\ }
  {}
  {}
\makeatother
\usepackage{siunitx}
\hypersetup{
    colorlinks=true,
    linkcolor=blue,
    filecolor=black,      
    urlcolor= MidnightBlue,
    citecolor=blue,
    pdftitle={Overleaf Example},
    pdfpagemode=FullScreen,
    }

\title{\LARGE \bf 
\SimName{}: Constructing Temporal Digital Twins of Plants with Gaussian Splats 
}
\author{Simeon Adebola$^{1}$, Shuangyu Xie$^{1}$, Chung Min Kim$^{1}$, Justin Kerr$^{1}$, Bart M. van Marrewijk$^{3}$,  \\ Mieke van Vlaardingen$^{3}$,  Tim van Daalen$^{3}$, E.N. van Loo$^{3}$, Jose Luis Susa Rincon$^{2}$, \\ Eugen Solowjow$^{2}$, Rick van de Zedde $^{3}$, and Ken Goldberg$^{1}$
\thanks{$^{1}$The AUTOLab at UC Berkeley (automation.berkeley.edu) {\tt\footnotesize \{simeon.adebola, goldberg\}@berkeley.edu}}%
\thanks{$^{2}$Siemens Research Lab, Berkeley, CA {\tt\footnotesize \{jose.susa\_rincon, eugen.solowjow\}@siemens.com}}%
\thanks{$^{3}$Netherlands Plant Eco-phenotyping Centre, Wageningen University and Research {\tt\footnotesize \{bart.vanmarrewijk, mieke.vanvlaardingen, tim.vandaalen, robert.vanloo, rick.vandezedde\}@wur.nl}}%
}

\begin{document}

\maketitle
\fancypagestyle{withfooter}{
  \renewcommand{\headrulewidth}{0pt}
  \fancyfoot[C]{\footnotesize Accepted to the Novel Approaches for Precision Agriculture and Forestry with Autonomous Robots IEEE ICRA Workshop - 2025}
}
\thispagestyle{withfooter}
\pagestyle{withfooter}

\begin{abstract}
Accurate temporal reconstructions of plant growth are essential for plant phenotyping and breeding, yet remain challenging due to complex geometries, occlusions, and non-rigid deformations of plants. We present a novel framework for building temporal digital twins of plants by combining 3D Gaussian Splatting with a robust sample alignment pipeline. Our method begins by reconstructing Gaussian Splats from multi-view camera data, then leverages a two-stage registration approach: coarse alignment through feature-based matching and Fast Global Registration, followed by fine alignment with Iterative Closest Point. This pipeline yields a consistent 4D model of plant development in discrete time steps. We evaluate the approach on data from the Netherlands Plant Eco-phenotyping Center, demonstrating detailed temporal reconstructions of Sequoia and Quinoa species. Videos and Images can be seen at \href{https://berkeleyautomation.github.io/GrowSplat/}{https://berkeleyautomation.github.io/GrowSplat/ }

\end{abstract}

\section{Introduction}

Plant phenotyping and breeding rely heavily on accurate measurements of plant structure over time. Traditional methodologies, which often involve manual or destructive sampling, can be both laborious and prone to human error. Recent advances in sensing and computation suggest the possibility of digital phenotyping platforms, enabling non-destructive, data-driven analyses of plant growth dynamics. In this context, \emph{digital twins}---virtual counterparts of physical entities---offer an attractive framework to model temporal changes in plant structures. By tracking and modeling a plant's morphology over time, breeders, researchers and practitioners could extract key traits (e.g., leaf angle, internode length) and study genotype-to-phenotype relationships under varied environmental conditions.

\begin{figure}
    \centering
    \includegraphics[width=\linewidth]{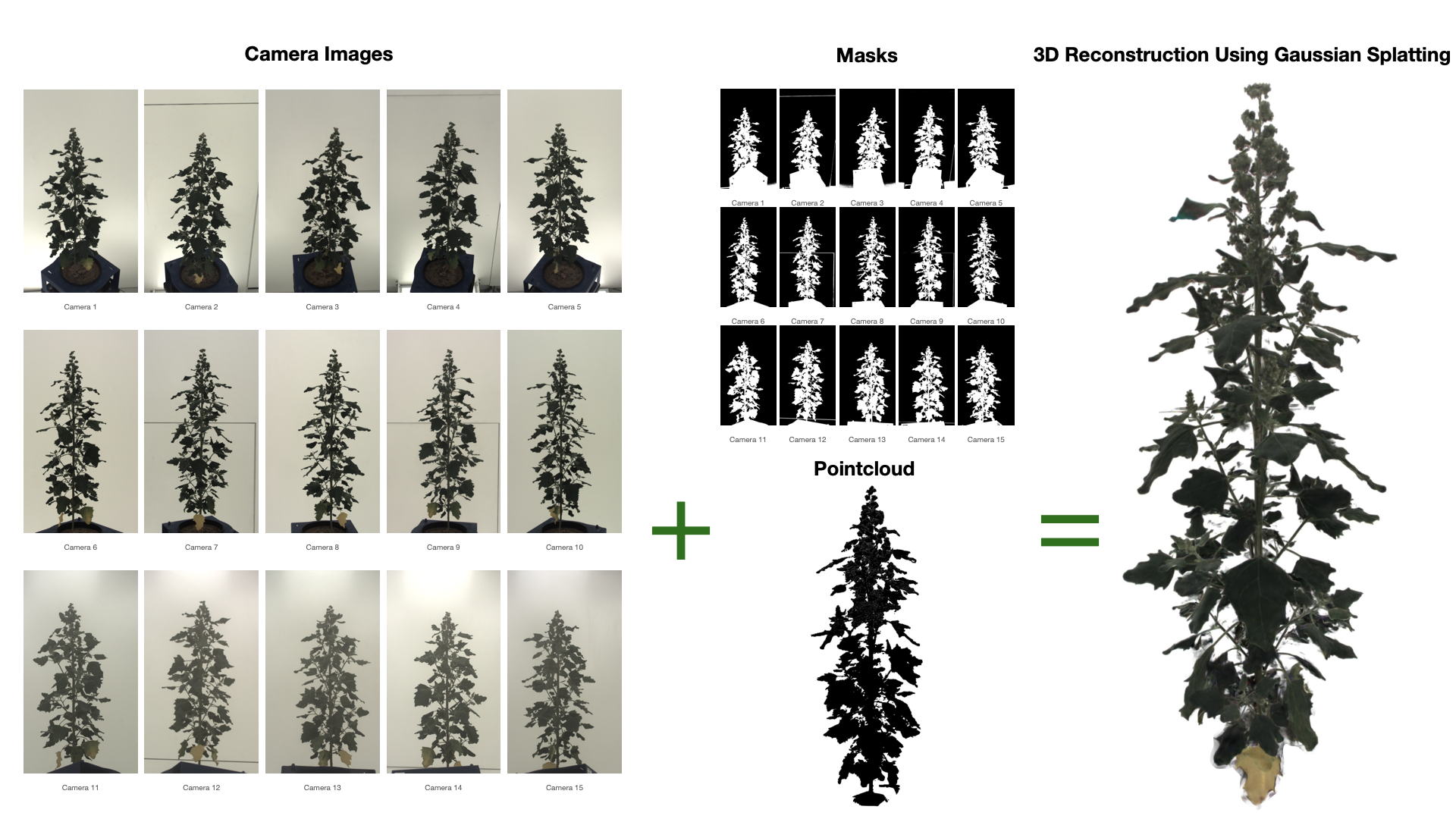}
    \caption{\textbf{\SimName{}} creates detailed 3D digital twins of plants with industrial-scale data and then constructs temporal digital twins over time. }
     \vspace{-.1in}
     \label{fig:splash}
\end{figure}

The Netherlands Plant Eco-phenotyping Center (NPEC) is a partnership between the Wageningen University \& Research and Utrecht University, with funding from the Netherlands Organisation for Scientific Research (NWO)\cite{about_npec}. NPEC is a leading phenotyping research organization \cite{10.3389/fbioe.2020.623705}  and seeks to provide large-scale automated phenotyping through high throughput and high resolution plant growth data. NPEC currently does this through seven research modules with advances facilities and tools: Ecotron, Plant-Microbe Interaction Phenotyping, Multi-Environment Climate Chambers, High-Throughput Phenotyping Climate Chambers, GreenHouse Phenotyping, Open-Field Phenotyping, and Data\cite{npec_phenotyping_modules}. Data collected from experiments using these modules allow for collaborations between both industry and academia partners. In this paper, we utilized 3D data collected from experiments in Module 5: GreenHouse Phenotyping (See Section \ref{section:system_overview}). 

Despite the promise of digital twins for plant science, building accurate and temporally consistent 4D plant models remains challenging. Plants exhibit complex geometries, occlusions, and non-rigid growth patterns that complicate the reconstruction process. Moreover, phenotyping studies typically involve substantial data volumes captured at multiple time points, compounding the computational burden. To address these challenges, we propose \SimName{}, an approach using \emph{Gaussian splats}---a continuous, scalable representation of point cloud data that naturally handles varying densities and irregularities in plant morphology. By converting  multi-view camera data provided by NPEC into this representation, we create temporal models of plant structures that can facilitate fine-grained, longitudinal trait analysis.

This paper presents registration and integration techniques, which align and merge sequential image scans into coherent temporal reconstructions, as well as strategies for handling both rigid and non-rigid deformations induced by plant growth. Ultimately, the proposed framework is a first step towards generating high-fidelity 4D models of plants across their growth cycles, opening new possibilities for data-rich phenotypic assessments and accelerating the breeding of more resilient crop varieties.

\section{Related Work}

\subsection{High-Throughput Industrial-Scale Plant Phenotyping}
\label{subsection:plantpheno}
In plant phenotyping, researchers measure characteristics such as height, width, mass, size, and area of various plant components such as fruits, leaves, and stems \cite{Esser2023field3D}. This data helps breeders and growers make informed choices to increase plant growth, reduce pests and diseases, and maintain a healthy plant ecosystem. 

High-Throughput phenotyping platforms have become more popular in recent years. These platforms vary in terms of the sensor types (e.g. RGB cameras, LiDAR, fluorescence sensors, ultrasonic, hyperspectral), phenotyping distance(e.g. aerial including satellite and unmanned aerial vehicles, ground-based/field, greenhouse or indoor),  traits measured(e.g. plant height, biomass, leaf area, water content, nitrogen content) , imaging techniques used, crops measured, data management systems and data processing approaches\cite{gill2022comprehensive, SONG2021633, 10.3389/fbioe.2020.623705}.

Traditional phenotyping approaches, including segmentation, relied on 2D information collected using sensors such as cameras and LiDAR \cite{ag_tase}. However, 2D phenotyping is often limited in the capture of leaf-level details due to occlusion \cite{gibbs2018plant}. Consequently, 3D methods are gaining preference for plant phenotyping due to reduced occlusion \cite{paulus2019measuring, harandi2023make}. Examples include using a transformer for 3D fruit shape completion \cite{magistri2024icra}, a neural network for the prediction of fruit shapes \cite{magistri2022ral-iros}, and 3D mapping, fruit shape and pose prediction with a mobile robot \cite{pan2023iros}. \cite{Lehnert_3DMTS} describes the use of a 3D camera array on a robot arm to identify the optimal position to capture the next image.

\subsection{3D Reconstruction}
\label{subsection:recon_lit}
Neural Radiance Fields (NeRFs) \cite{mildenhall2020nerf} were introduced in 2020. NeRFs work by optimizing a volumetric representation using posed images~\cite{mildenhall2020nerf, nerfstudio}. Several fields including robot grasping \cite{pmlr-v164-ichnowski22a, pmlr-v205-kerr23a, lerftogo2023, shen2023F3RM} have used NeRFs. A more recent technique called three-dimensional (3D) Gaussian Splatting (3DGS), achieves scene reconstruction and novel view synthesis comparable to NeRFs \cite{kerbl3Dgaussians} and has been shown to have a faster rendering time \cite{kerbl3Dgaussians, he2024nerfs}. As a result, researchers that use NeRFs have also shown updated results with 3DGS including in grasping \cite{zheng2024gaussiangrasper3dlanguagegaussian}.

\subsection{4D Plant Modeling}
Growth tracking and temporal modeling of plants present unique challenges due to the non-rigid deformations inherent in biological growth processes. \cite{li2013analyzing} proposed a spatio-temporal reconstruction framework specifically designed for aligning and modeling of plants 3D point clouds. Their method addresses the challenging problem of tracking plant growth by incorporating both spatial and temporal constraints with annotated alignment.

Researchers have approached 4D plant modeling primarily as a registration problem. Building on this foundation, other researchers have explored alternative approaches to temporal plant registration. Chebrolu et al. \cite{chebrolu2020spatio} have utilized skeletal structures as geometric priors to guide the registration process, while others have employed learning-based methods to predict growth patterns. Paulus et al. \cite{paulus2019measuring} utilized morphological features for phenotyping applications, whereas Harandi et al. \cite{harandi2023make} focused on developing processing techniques specifically for 3D plant phenotyping.

\section{System Overview}
\label{section:system_overview}
\subsection{The \Setup{} Setup}
\begin{figure}
    \centering
    \includegraphics[width=\linewidth]{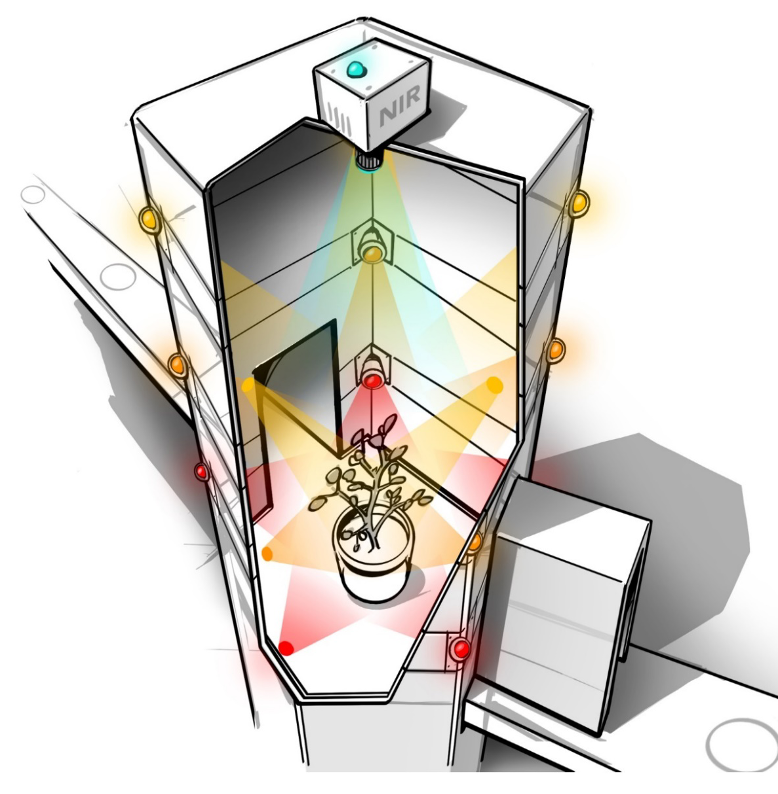}

    \makebox[\linewidth]{%
        \includegraphics[height=4cm, width=5.3cm]{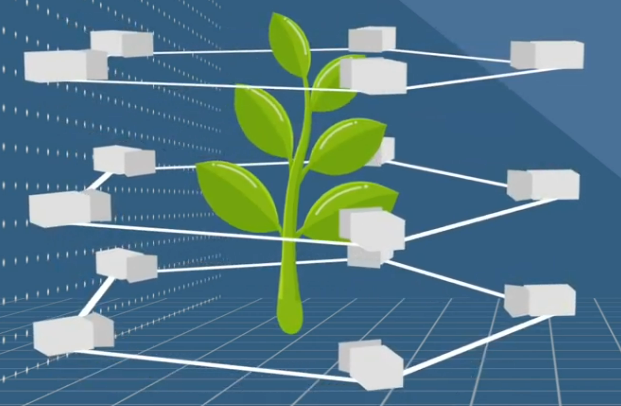}
        \includegraphics[height=4cm]{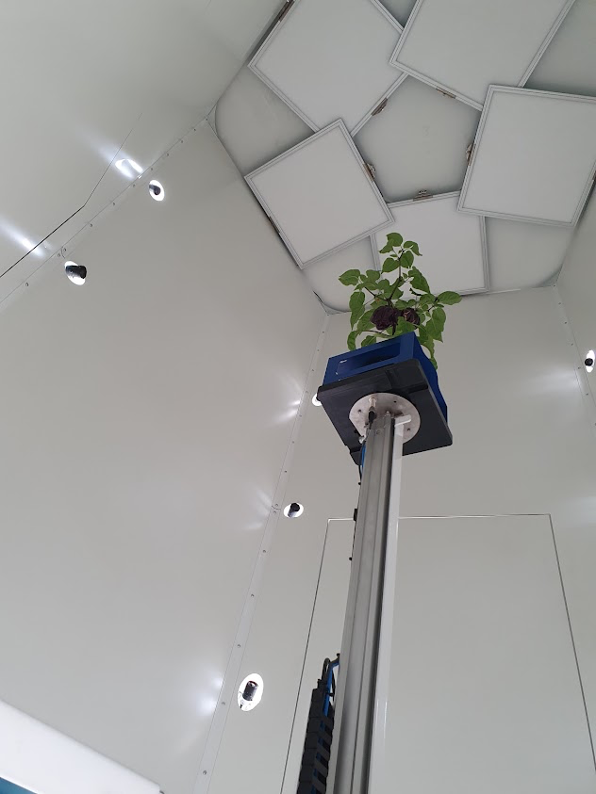}
    }
    \caption{\textbf{\Setup{}} is an indoor system for plant phenotyping that constists of 15 calibrated static cameras. Plants are moved into the Maxi-Marvin using a conveyor belt and 15 images are taken. The system can be used for large plants up to a height of 70cm. }
     \label{fig:maxi-marvin}
\end{figure}

The NPEC data collection system is known as the \Setup{} \cite{maxi_marvin}. \Setup{} consists of 15 static cameras arranged in three layers of five cameras each. The cameras are calibrated and maintain the same position while \Setup{} is integrated into a conveyor belt system at NPEC thus allowing  multiple plants to be quickly imaged. Figure ~\ref{fig:maxi-marvin} shows the \Setup{} setup.  When a plant is moved into the \Setup{}, each camera takes an image. These 15 images together with the calibrated poses for each camera is the input. 

For each camera in the \Setup, once calibration is done, we are provided with 3D pose parameters (rotation angles and translation vector), camera intrinsics, and internal camera parameters ( focal length, radial distortion coefficient, image width and image height, image center coordinates, and scale factors). \Setup{} uses the division model as its distortion model instead of the polynomial model. As a result, the internal camera parameters only have one value for the  radial distortion coefficient  whereas the polynomial model has five values: three for radial distortion and two for decentering distortions \cite{calibrate_cameras_nodate}.

\subsection{Data Preprocessing for NerfStudio}
For 3D reconstruction we use Nerfstudio\cite{nerfstudio}. Using the pose parameters, camera intrinsics and internal camera parameters, we are able to  generate the required poses for each image from each camera. However, Nerfstudio uses six distortion parameters\cite{nerfstudiocamerascamera_utils_nodate} so we convert the single coefficient $\kappa$ into four values:  K1 =  $-\kappa/\sqrt{w^2+h^2}$, K2 = $(-\kappa/\sqrt{w^2+h^2})^2$, P1 = 0.0, P2 = 0.0 where w is the image width and h is the image height. The last two distortion parameters (K3 and K4) are set to 0.0 by default.

\section{Problem Statement}

Modeling plant growth over time requires capturing and aligning sequential 3D representations of plants. Given a sequence of point clouds obtained at distinct durations (approximately 1 day), the objective is to register plant models accurately in a common reference frame considering both rigid and non-rigid deformations due to plant growth.

\subsection{Assumptions}
To formulate the problem, we make the following assumptions:
\begin{itemize}
    \item[a.1] The plant is observed on days $t_0, t_1, t_2 ...$ where $Avg(\Delta t) < 3$ days. 
    \item[a.2] Plant scans are performed under similar environmental conditions with minimal external disturbances. 
\end{itemize}
These assumptions are met by the system as referenced in the prior section.

\subsection{\textit{Notation}}
Variables are defined as follows:

\begin{itemize}
    \item $\mathcal{K} = \{t_{1}, ..., t_{K}\}$: Discrete time index set which represents a sequence of time indexes that corresponds to the plant growth cycle, where $t_{k} \in \mathcal{K}$ is the $k$-th sampling day.
    \item $\mathcal{W}$: global plant frame. A right-handed 3D Euclidean coordinate system such that the coordinate origin aligns with the bottom part of the visible stem and the z-axis aligns with the plant growing direction (e.g. the main stem direction). The rotation for world frame is determined by the camera system of \Setup{}. 

    \item $P_{t_k} \subset \mathbb{R}^3$: 3D point cloud at time step $k$ in the local sensor frame.
    \item $\mathcal{P} = \{ P_{t_k}: t_k\in \mathcal{K}\}$: Sequence of 3D point clouds of a plant.
    \item $P_{\text{ref}}$: A chosen reference point cloud,  serving as the fixed coordinate frame for registration, typically the observation from the previous time index in the world frame $\mathcal{W}$.
    \item $\mathcal{T}_{t_k}$: Transformation function aligning $P_{t_k}$ to a reference frame $P_{\text{ref}}$.
\end{itemize}

\subsection{Transformation Model}

Plant growth involves both rigid transformations, such as translations and rotations caused by sensor movement, and non-rigid deformations, including leaf expansion and stem bending driven by biological growth processes. 
\begin{equation}
    \mathcal{T}_{t_k}(P_{t_k}) = T_{t_k} P_{t_k} + \mathbf{d}_{t_k}
\end{equation}
Specifically, we denote $T_{t_k} \in SE(3)$ as global rigid alignment,
For point cloud $P_{t_k}$ with $n_{t_k}$ points,
$\mathbf{d}_{t_k} \in \mathbb{R}^{n_{t_k}}$ is a non-rigid deformation field modeling growth-induced shape changes.

\begin{Pro}[Temporal Plant Registration] \label{pro:plant_modeling}
Given a sequence of plants model, estimating a transformations function $\mathcal{T}_{1:K}$ that aligns each $P_{t_k}$ to a reference frame $P_{\text{ref}}$, such that:
\begin{align}
    \min_{T_{1:K}, \mathbf{d}_{1:K}} \quad & \sum^{K}_{k=1} \mathcal{L}(P_{t_k}, P_{\text{ref}})  \\
    \textrm{s.t.} \quad &  \| T_{t_k}T^{-1}_{t_{k-1}} \|_{\Sigma} \le \alpha  , \quad \forall k \in \mathcal{K} \label{eq:t-cond} \\
     &  \mathbf{d}^\mathsf{T}_{t_k} \mathbf{d}_{t_k} \le  \beta,
\end{align}
where $\mathcal{L}$ is a registration loss function that quantifies the misalignment between the transformed point cloud and the reference,  $\alpha$ and $\beta$ are the growth parameters determined by specific plant. 
\end{Pro}

\section{Method}

We propose a 4D plant modeling method that first reconstructs 3D Gaussian splats from the \textit{NPEC \Setup{} system}, then performs temporal alignment by solving Problem~\ref{pro:plant_modeling}, and finally employs a holistic rendering of the plant growth model to integrate the reconstructed gsplats into a coherent 4D representation.

\begin{figure*}[ht!]
    \centering
    \includegraphics[width=\textwidth]{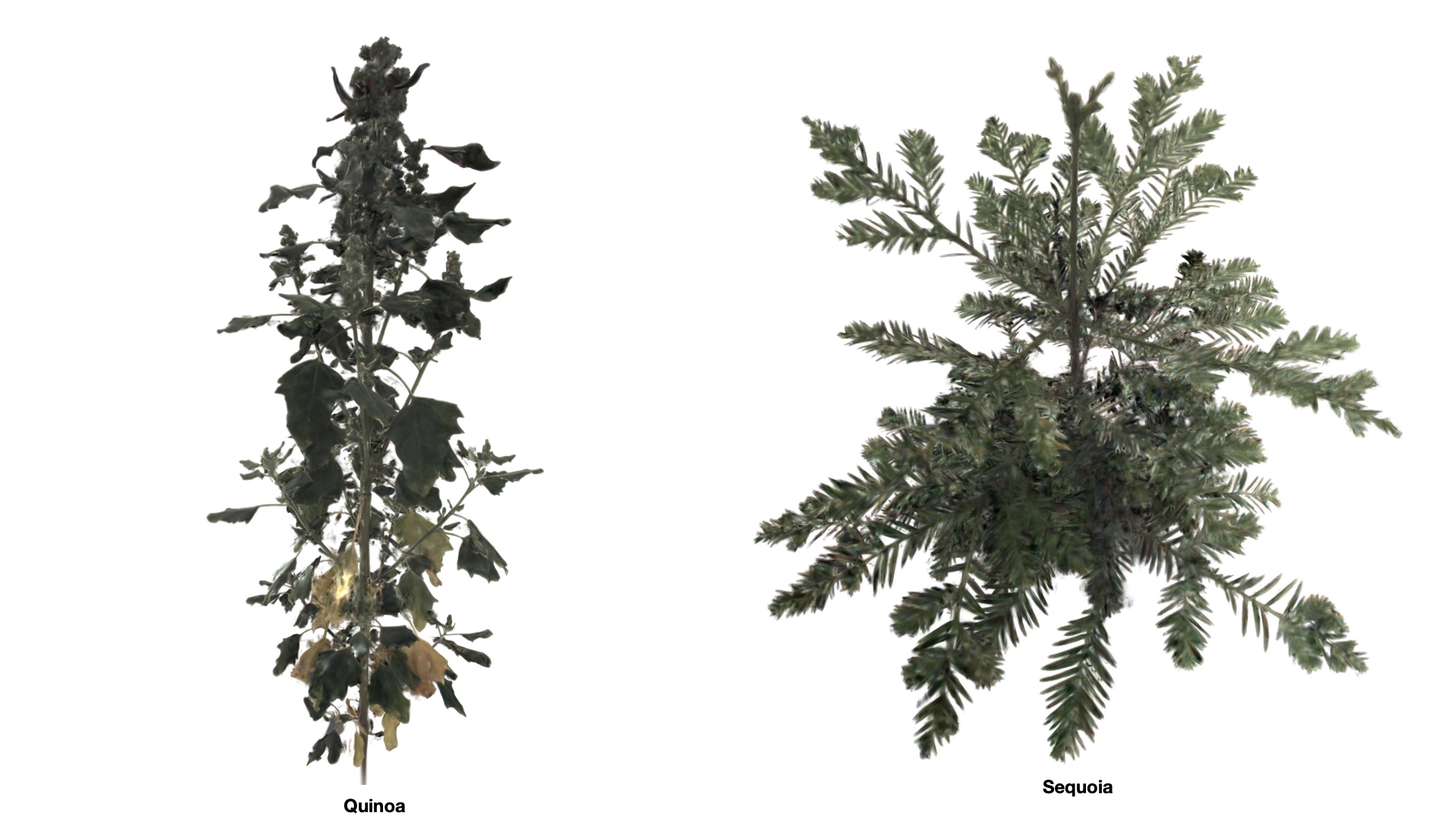}
    \caption[width=\textwidth]{\textbf{\SimName{} Digital Twins:} Presented here are a side view for two different plants.  Each column shows an RGB view of the 3D model.}
    \vspace{0.3cm}
\label{fig:nerfacto}
\end{figure*}

\begin{figure*}[ht!]
    \centering
    \includegraphics[width=\textwidth]{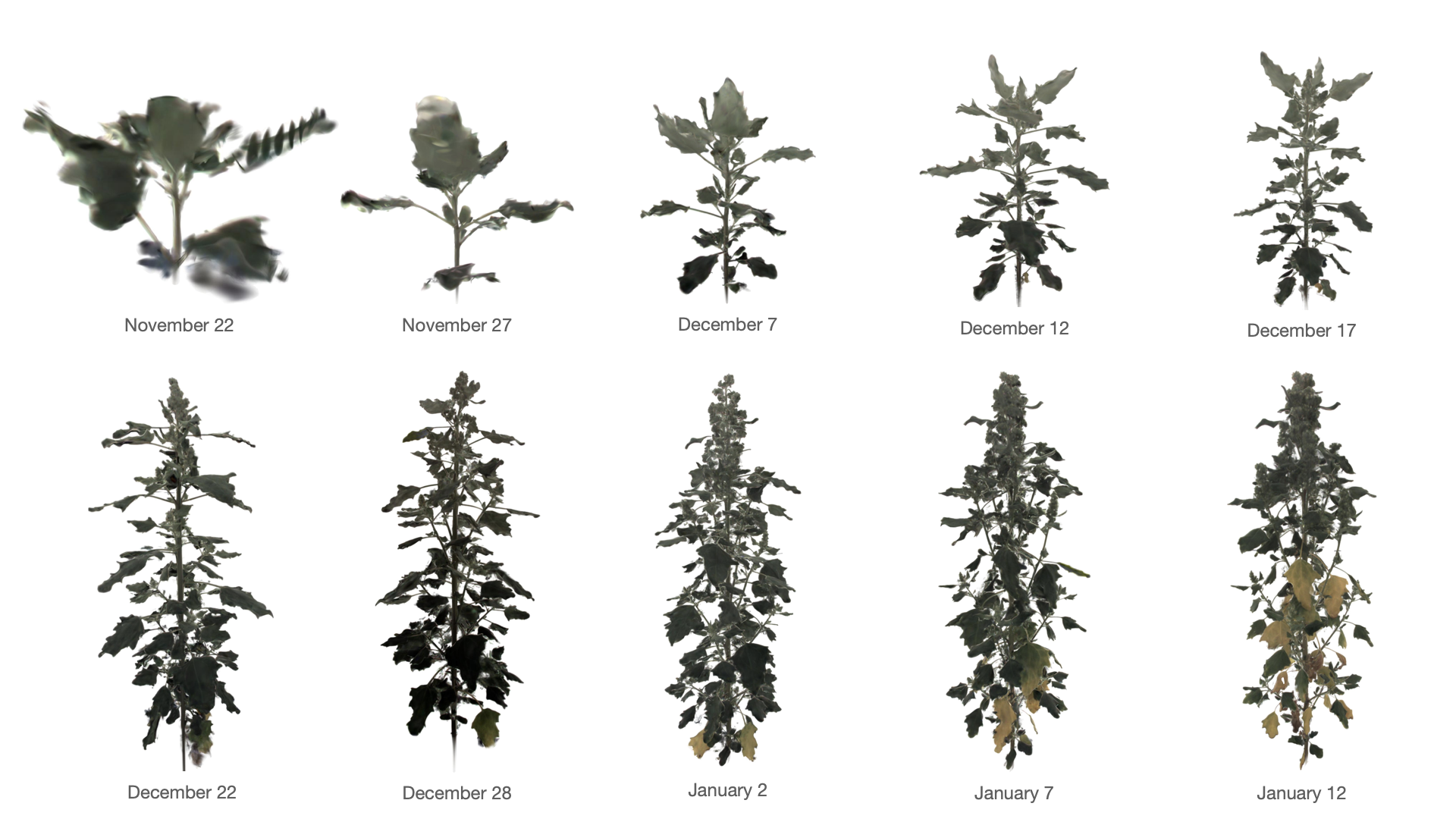}
    \caption[width=\textwidth]{\textbf{\SimName{} Digital Twins:} Presented here are ten reconstructions of Quinoa across a period. The plant's growth is seen over time.}
    \vspace{0.3cm}
\label{fig:quinoa}
\end{figure*}
\begin{figure*}[ht!]
    \centering
    \includegraphics[width=\textwidth]{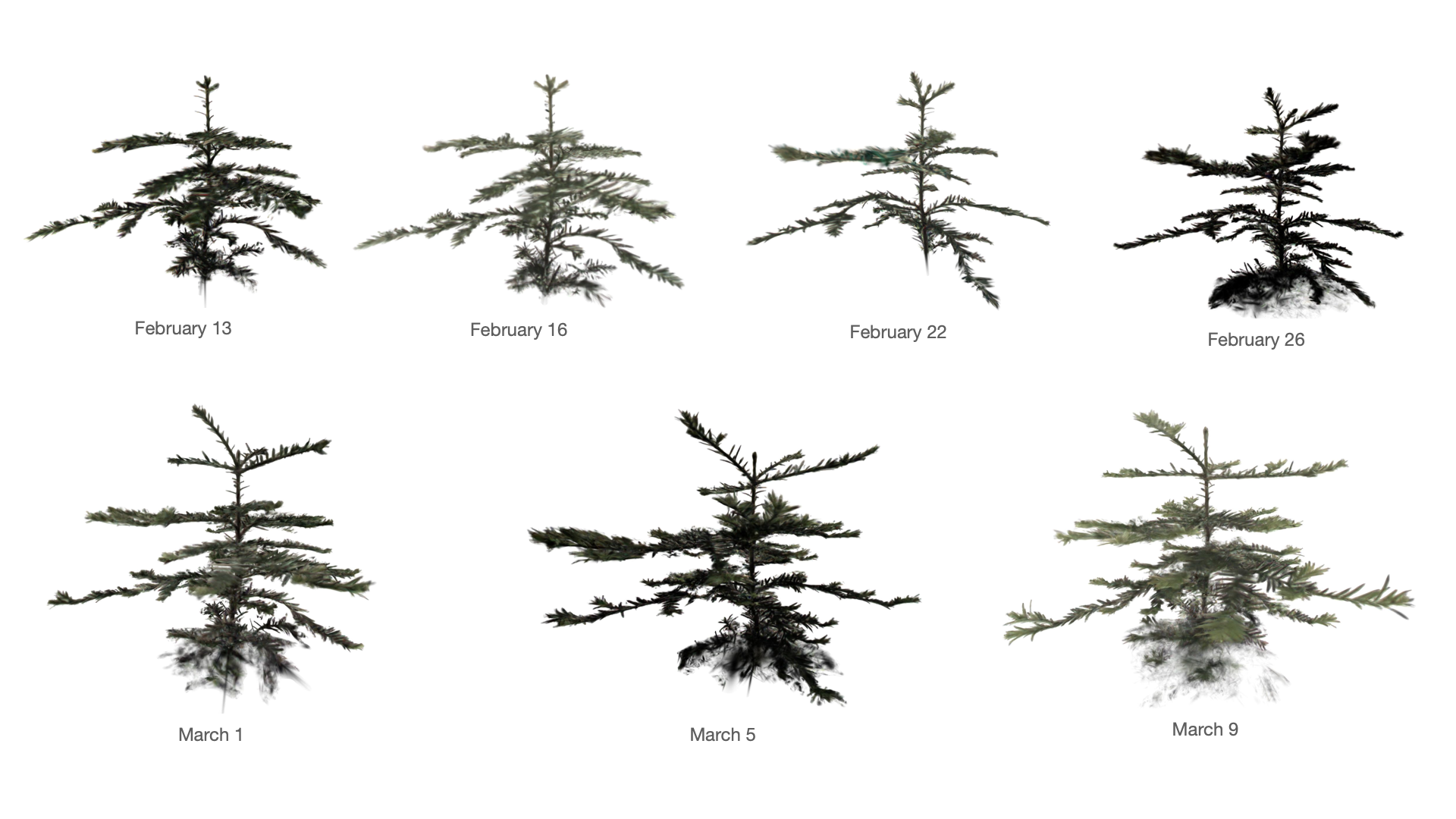}
    \caption[width=\textwidth]{\textbf{\SimName{} Digital Twins:} Presented here are seven reconstructions of Sequoia across a period. The plant's growth is seen over time.}
    \vspace{0.3cm}
\label{fig:sequoia}
\end{figure*}

\subsection{3D Gaussian Splats}
We begin by generating high-quality 3D plants reconstructions at each time step using 3D Gaussian splatting \cite{kerbl3Dgaussians}. Unlike NeRFs which require lengthy training times, 3DGS provides comparable reconstruction quality with significantly faster training speeds (it takes 3 seconds for GS-based methods to train the first 100 iterations versus 6 seconds for NeRF-based methods \cite{he2024nerfs})  making it ideal for industrial-scale plant phenotyping applications. We train our 3DGS models using Splatfacto\cite{nerfstudio, ye2023mathematical}. Specifically we use Splatfacto-MCMC which uses the  3D Gaussian Splatting as Markov Chain Monte Carlo\cite{kheradmand20243d} strategy to train the Gaussian splats. Since we only have fifteen images, we use segmentation masks that segment the plants from the rest of the image. When training the GS model, we do not compute standard loss functions on pixels lying outside these masks, and we utilize a loss minimizing total rendered opacity. We also utilize initial pointclouds provided by NPEC. So our reconstruction is aided by using both the initial pointclouds and the segmentation masks. We also introduce a lighting factor as some of the plant images from the \Setup{} are darker which affect the reconstruction quality. Splatfacto models are trained using on a
computer with an AMD Ryzen 7 7700x 8-core processor × 16 processor, an NVIDIA GeForce RTX 4090 GPU and 64.0GB Memory.

Some examples of 3D gaussian splats can be found in Fig ~\ref{fig:nerfacto}, Fig ~\ref{fig:quinoa}, and Fig ~\ref{fig:sequoia}

\subsection{Plant registration}
To solve Prob.~\ref{pro:plant_modeling} from the sequence of time-indexed Gaussian splats, we propose the algorithm framework as follows:
\subsubsection{Point Cloud Initialization and Preprocessing}
Gaussian splats provide a dense point cloud with an associated uncertainty model; however, using the raw data directly leads to excessive computational overhead and introduces outliers stemming from the uncertainty. To address these challenges, we propose a point cloud downsampling pipeline for Gaussian splats that applies three principal filtering conditions with Gaussian splats properties to ensure physically valid and well-conditioned data. First, we discard splats with excessively large or small scales by inspecting their log-scale range, thus preventing skewed modeling. Second, we compute the scale ratio to remove overly elongated splats that do not faithfully represent plant geometry. Lastly, we validate the rotation parameters by checking the norm of the associated quaternion to reduce numerical errors. After this filtering process, we estimate surface normals to support the computation of Fast Point Feature Histograms (FPFH)\cite{rusu2009fast}. These FPFH descriptors capture local geometry and serve as key features in subsequent alignment steps.

\subsubsection{Global Registration (Coarse Alignment)}
For an initial, coarse alignment of point clouds across different time steps, we adopt a feature-based matching approach inspired by Fast Global Registration (FGR)~\cite{Zhou2016}. Specifically, we first compute FPFH for each point in both the reference point cloud \( P_{\mathrm{ref}} \) and the temporal point cloud \( P_{t_k} \). The FPFH descriptors encode local geometric properties (e.g., curvature, normal variation) that are invariant to rigid transformations, facilitating the discovery of putative correspondences between the two point clouds.We then utilize a RANSAC-based scheme for robust outlier rejection. In each iteration of RANSAC, a minimal set of correspondences is randomly sampled, and an initial rigid transformation is estimated. The transform is evaluated against the entire set of FPFH correspondences to identify and reject outliers. We set the convergence condition to be \eqref{eq:t-cond} or the maximum number of iterations reach the threshold. The resulting transformation \( T_{t_k} \) approximately aligns \( P_{t_k} \) to \( P_{\mathrm{ref}} \). Once we have a candidate set of correspondences that survive RANSAC, we refine them using the optimization procedure proposed in FGR, which employs a robust objective to handle remaining outliers more gracefully than pure least-squares approaches. By combining FPFH-based matching, RANSAC filtering, and FGR optimization, the final output of this global registration step is a coarse but sufficiently accurate alignment of \( P_{t_k} \) to \( P_{\mathrm{ref}} \). This ensures a robust starting configuration for subsequent local (fine) registration.

\subsubsection{Fine Registration (Local Alignment)}
Following coarse alignment, we refine each temporally acquired point cloud using standard Iterative Closest Point (ICP), which iteratively refines the rigid transformation by minimizing the Euclidean distance between matched points. To further improve alignment, we employ Colored ICP, which incorporates photometric consistency by minimizing color discrepancies between corresponding points. This color term helps disambiguate similar geometric features, particularly in organic structures with subtle variations.

\subsection{View Rendering}
\begin{figure}
    \centering
    \includegraphics[width=\linewidth]{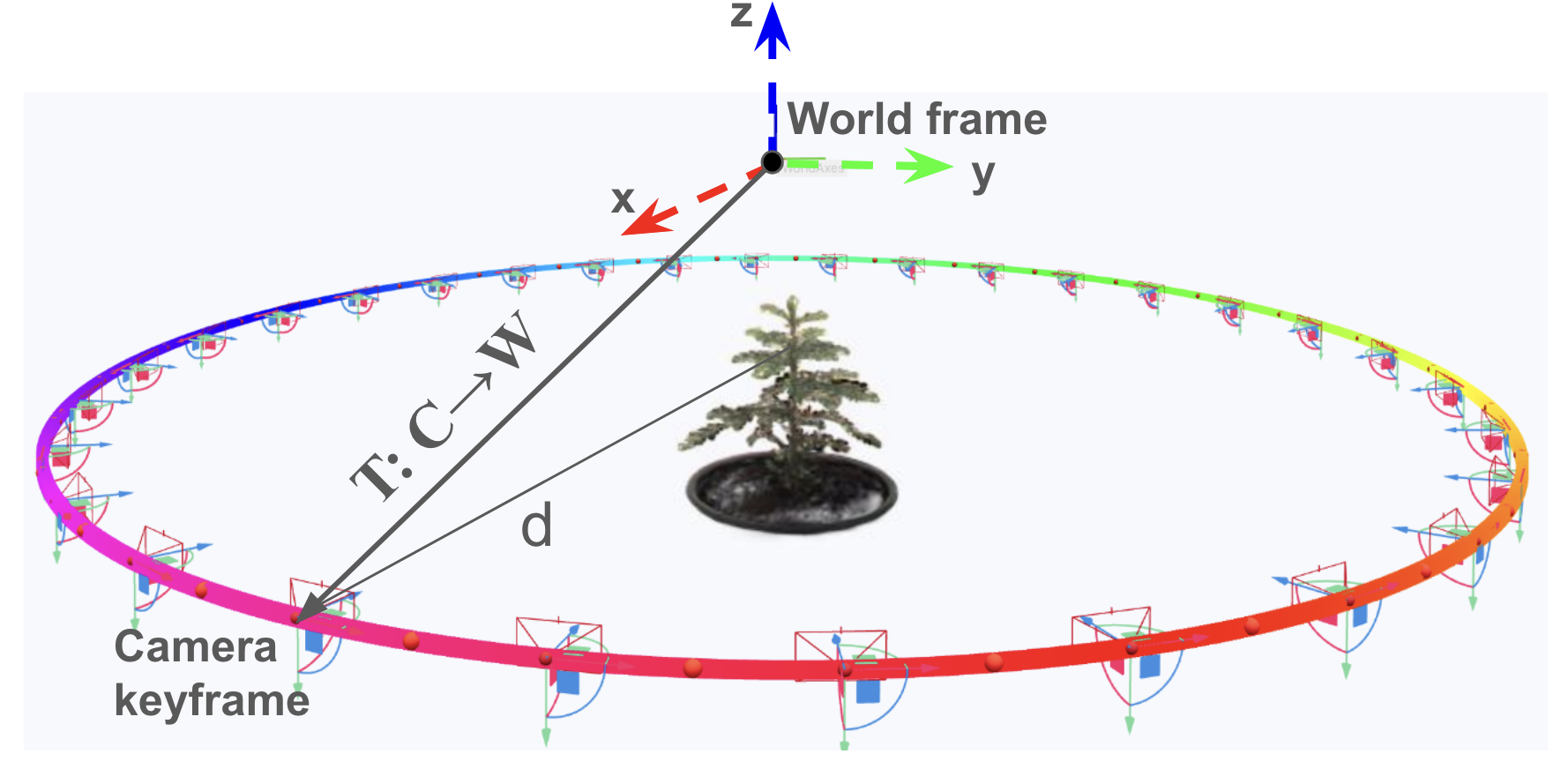}
    \caption{Camera Keyframes Around Reconstruction}
    \label{fig:rotate}
\end{figure}
We use the registered plant frame and the reconstructed Gaussian splats to render rotating viewing angles, as shown in Fig.~\ref{fig:rotate}. The final output is a temporally consistent sequence of 3D point clouds and video that best captures the growth dynamics of the plant.

\section{Experiments}
We evaluated \SimName{} with plant species Sequoia and Quinoa that were captured by NPEC \Setup{} system with longest time series with 55 time points over 76 days.

\subsection{Dataset}
Table~\ref{tab:leaf_metrics} provides a high-level summary of the datasets used in our experiments collected with multi-view cameras. Each entry corresponds to a different plant species, along with its respective observation period (indicated under Duration), the total number of temporal snapshots (denoted by \( |\mathcal{K}| \)), and the average number of days between each snapshots (denoted by $Avg(\Delta t)$). The Sequoia dataset spans from February~13, 2024, to May~24, 2024, with 40 time steps, whereas the Quinoa dataset covers the period from October~31, 2022, to January~15, 2023, totaling 55 time steps.

\begin{table}[t]
\footnotesize
\centering
\begin{tabular}{lccc} %
\toprule
\textbf{Plant} & \textbf{Duration} & $|\mathcal{K}|$ & $Avg(\Delta t)$\\ 
\midrule
Sequoia  & 2024-02-13 to 2024-05-24  &  40 & 2.4\\
Quinoa  & 2022-10-31 to 2023-01-15  & 55 & 1.4\\
\bottomrule
\end{tabular}
    \caption{\textbf{Dataset Overview}}
\vspace{-1em}
\label{tab:leaf_metrics}
\end{table}

\subsection{Result}
More videos can be found on the website. Fig ~\ref{fig:quinoa} and Fig ~\ref{fig:sequoia} show plant reconstructions over time. 

\begin{figure}
    \centering
    \includegraphics[width=\linewidth]{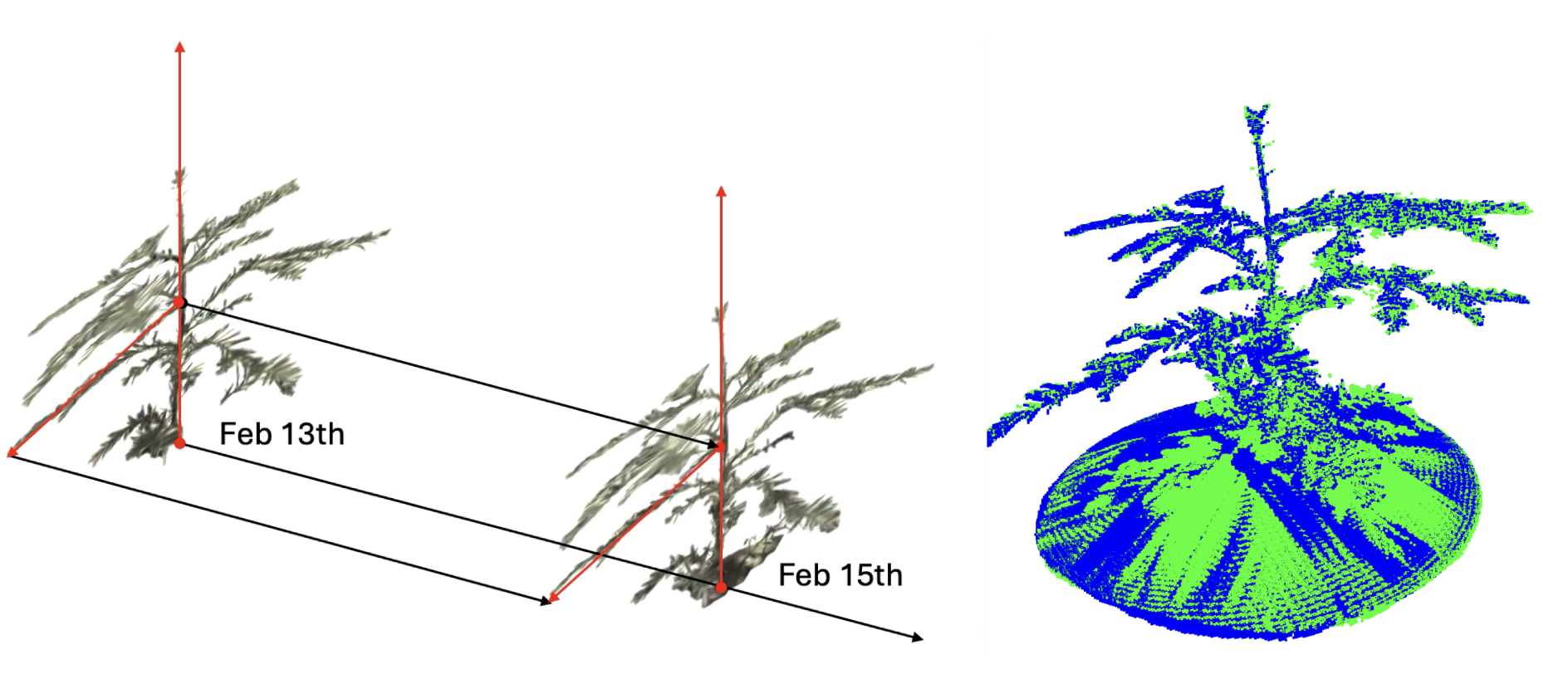}
    \caption{An example registration between two timestamps of Sequoia. }
    \label{fig:enter-label}
\end{figure}

\begin{figure}
    \centering
    \includegraphics[width=\linewidth]{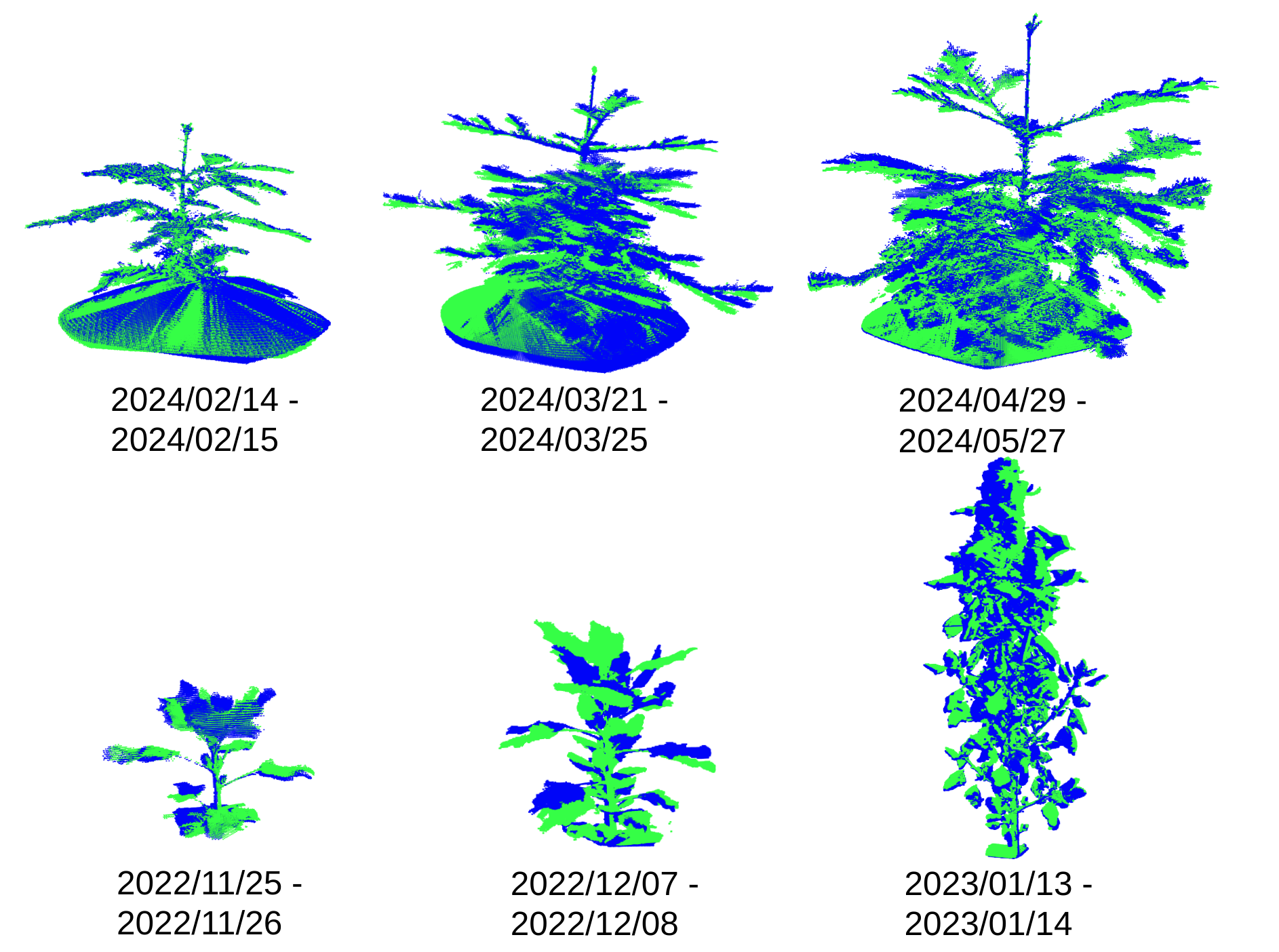}
    \caption{Top row: registration result for Sequoia. Bottom row: registration result for Quinoa. We use the blue to indicate the point cloud in the earlier date and the green indicate the point cloud in the later date.  }
    \label{fig:enter-label}
\end{figure}

\section{Limitations and Future Work}
For biologically plausible growth patterns and temporal consistency, in the future we will implement several constraints on biological priors such as monotonic leaf area growth and structural consistency of stems and branches. These constraints can reduce physically impossible transformations that might result from noise or registration errors. Meanwhile, we can interpolate between frames to estimate the plant morphology in the unobserved time stamp. 

We will also carry out quantitative evaluations of \SimName{} to further show its effectiveness and usefulness for desired downstream tasks.  

We are also exploring biomass estimation using ground truth measurements for the plants and point clouds from the reconstruction.

\section{Conclusion}
We present \SimName{}, an implemented system for autonomously creating detailed 3D industrial scale digital twins of plants. These digital twins are augmented with segmentation information, leaf detections, and physical properties analyzed from the model which could facilitate more efficient plant phenotyping at scale.

\bibliographystyle{IEEEtran}
\bibliography{IEEEabrv,references}

\end{document}